\documentclass[10pt]{article} 

\usepackage[preprint]{rlj} 

\usepackage{amssymb}            
\usepackage{mathtools}          
\usepackage{mathrsfs}           
\usepackage{graphicx}           
\usepackage{subcaption}         
\usepackage[space]{grffile}     
\usepackage{url}                
\usepackage{lipsum}             
\usepackage{booktabs}
\usepackage{makecell}
\newenvironment{DIFnomarkup}{}{}
\usepackage[table]{xcolor}
\definecolor{oddrowcolor}{gray}{1.0}
\definecolor{evenrowcolor}{gray}{0.9}
\usepackage{float}

\title{When to Plan: Learning to Select Between Reactive Control and Deliberative Planning}

\setrunningtitle{When to Plan: Learning to Select Between Reactive Control and Deliberative Planning}

\author{Adam Labiosa\textsuperscript{1}, Josiah P. Hanna\textsuperscript{1}}

\emails{labiosa@wisc.edu, jphanna@cs.wisc.edu}

\affiliations{
$^{1}$\textbf{Department of Computer Sciences, University of Wisconsin--Madison, USA}
}

\contribution{
We introduce a reinforcement learning approach to learning a meta-policy for selecting between a fast, reactive policy, which may perform poorly out of its training distribution, and a planner, which requires time but reliably produces near-optimal actions.
}
{
Prior work has used RL to minimize compute, either for a single decision by selecting internal computations \citep{callaway_learning_2018} or across an episode by using action repetitions \citep{orenstein_toward_2025}. Our work differs by addressing the meta-reasoning problem as needing to make a sequence of meta-level selections between a reactive policy and a planner across a full task episode. Related ideas have been studied in cognitive science \citep{keramati_speedaccuracy_2011, kool_planning_2018}, but these works aim to describe human behavior rather than provide mechanisms for artificial agents.
}

\contribution{
    We show that our method produces meta-policies that balance deliberate planning and using a reactive policy as a function of environment characteristics such as planning delay, reactive policy competence, and environment stochasticity.
    }
    {
     We conduct an empirical study which shows our method obtains higher return than baselines that always commit to one type of decision-rule (\hyperref[tab:main_results]{Table \ref{tab:main_results}}). By varying planning cost, reactive policy competence, and task stochasticity, we analyze how each factor changes the balance of reactive action compared to deliberative planning (\hyperref[fig:env_experiment]{Figure \ref{fig:env_experiment}}).
    }
    
\contribution{
    We show that an uncertainty-conditioned meta-policy can track reactive policy competence even if the reactive policy improves over time, shifting toward more reactive control as reactive competence increases.
    }
    {
    We conduct an experiment that uses behavior cloning on planner rollouts to fine-tune the reactive policy during training (\hyperref[sec:joint_training]{Section \ref{sec:joint_training}}, \hyperref[fig:joint_training_plot]{Figure \ref{fig:joint_training_plot}}). The key result is that uncertainty-conditioned observations are sufficient for the meta-policy to learn to call the reactive policy more as its training distribution widens.
    }


\summary{
It has long been recognized that humans have the ability to switch between fast, reactive decision-making and slower, deliberative planning.
In this paper, we study the question of how to learn this ability, known as meta-reasoning, in artificial agents. 
We model reactive decision-making as a policy that directly maps state observations to actions. 
Such policies can be trained with reinforcement learning (RL) or imitation learning, but may generalize poorly outside of their training distribution. 
Alternatively, model-based decision-time planning is more likely to produce good actions across a broader set of states but requires additional computation time, which delays acting. 
In this work, we introduce an RL method for training a meta-reasoning policy that allocates computation by conditioning on a reactive-policy uncertainty score.
This score enables it to predict when the reactive policy is likely to perform poorly and when planning is needed. 
We conduct an empirical study on motion planning and navigation environments, showing that this design enables the meta-reasoning policy to learn when the reactive policy provides a good-enough action versus when decision-time planning is needed. 
Additionally, we show that our design enables the meta-agent to shift toward fully reactive control as the reactive policy improves.
}

\begin{document}

\makeCover  
\maketitle  


\begin{abstract}
It has long been recognized that humans have the ability to switch between fast, reactive decision-making and slower, deliberative planning.
In this paper, we study the question of how to learn this ability, known as meta-reasoning, in artificial agents. 
We model reactive decision-making as a policy that directly maps state observations to actions. 
Such policies can be trained with reinforcement learning (RL) or imitation learning, but may generalize poorly outside of their training distribution. 
Alternatively, model-based decision-time planning is more likely to produce good actions across a broader set of states but requires additional computation time, which delays acting. 
In this work, we introduce an RL method for training a meta-reasoning policy that allocates computation by conditioning on a reactive-policy uncertainty score.
This score enables it to predict when the reactive policy is likely to perform poorly and when planning is needed. 
We conduct an empirical study on motion planning and navigation environments, showing that this design enables the meta-reasoning policy to learn when the reactive policy provides a good-enough action versus when decision-time planning is needed. 
Additionally, we show that our design enables the meta-agent to shift toward fully reactive control as the reactive policy improves.
\end{abstract}

\section{Introduction}

Humans naturally allocate cognitive effort depending on the situation. For example, drivers often navigate familiar routes with minimal cognitive effort, whereas in unfamiliar environments they shift from following intuitive reactions to deliberate decision-making \citep{charlton_driving_2013}. 
Acting reactively as opposed to thinking before acting has been termed System 1 and System 2 thinking in cognitive science \citep{kahneman2011thinking, evans1984heuristic}, and a similar distinction can be seen in AI systems. 
Reactive, pattern-matching agents trained with model-free RL or imitation learning enable quick decision-making but may generalize poorly to states outside of their training distribution, whereas agents that always invoke a decision-time planner produce high-quality actions but must spend time creating a plan before acting. 
Most current systems commit to a fixed strategy at test time, like, for example, always using model-free reactive actions \citep{bansal_chauffeurnet_2018, haarnoja_learning_2023}, always planning before acting \citep{xia_survey_2024}, or always reasoning before responding \citep{openai_openai_2024, deepseek-ai_deepseek-r1_2025}.

The goal of this work is to enable \emph{adaptive computation} by learning \textit{when} to invoke more computationally-intensive planning routines as opposed to taking a reactive action. 
Our work contributes to the area of AI research known as meta-reasoning \citep{horvitz_computation_nodate, russell_principles_1991}, the study of how an agent should allocate its computation, and specifically the use of RL to learn policies for meta-cognition.

We study the meta-reasoning problem in a setting with two primary action-generation mechanisms: a reactive policy and a decision-time planner. 
The reactive policy produces an action from a single forward pass of a neural network but only performs well within a limited training distribution. 
In contrast, the planner produces a good sequence of actions from any environment state, albeit at the cost of additional computation compared to the reactive policy. 
While learned world models used for planning are similarly limited to their training distribution, prior work has shown that they generalize more robustly than reactive policies \citep{young_benefits_2023}.
Planning is therefore most valuable in states outside of the reactive policy's training distribution.
These observations motivate the need to arbitrate between a distribution-constrained reactive policy and a planner that is effective across the full state space.

Within this problem setting, we ask the following question:
\begin{center}
    \emph{How can an RL agent learn to select between a distribution-constrained, reactive policy and a slower decision-time planning routine?}
\end{center}

To answer this question, we develop a method to train a meta-policy that selects between reactive action and variable-depth planning to maximize task return under a time-based cost. Critically, the meta-policy is conditioned on uncertainty signals from the reactive policy, which allow it to estimate reactive competence in the current state.

We evaluate our approach in a suite of motion planning and navigation tasks. First, we show our adaptive meta-policy takes less time to reach goal states in comparison to fixed-compute baselines, including always react and always plan. Second, we ablate key design choices of the meta-policy observation space and show that reactive uncertainty and observation history are the most critical components for effective compute allocation. Third, we demonstrate that the meta-policy learns environment characteristic-dependent behavior by varying the reactive competence, planning cost, and environment stochasticity. Lastly, we apply our method to a joint-training setting where the reactive policy improves alongside the meta-policy and show that uncertainty-conditioned observations enable the meta-policy to track increasing reactive competence and shift toward fully reactive control.

In summary, our work makes the following contributions:
\begin{enumerate}
    \item We introduce a novel reinforcement learning approach to learning a meta-policy for selecting between a fast, reactive policy, which may perform poorly out of its training distribution, and a planner, which requires time but reliably produces near-optimal actions.
    \item We show that our method produces meta-policies that balance deliberate planning and using a reactive policy as a function of environment characteristics such as planning delay, reactive policy competence, and environment stochasticity.
    \item We show that an uncertainty-conditioned meta-policy can track reactive policy competence even if the reactive policy improves over time, shifting toward more reactive control as reactive competence increases.
\end{enumerate}

\section{Markov Decision Processes}
\label{sec:background}

We consider a Markov decision process (MDP) defined by the tuple
$\mathcal{M} = (\mathcal{S}, \mathcal{A}, P, r, \gamma)$, where
$\mathcal{S}$ is the state space, $\mathcal{A}$ is the action space, $P(s' \mid s, a)$ denotes the transition dynamics, $r(s, a)$ is the reward function, and $\gamma \in [0,1)$ is the discount factor. At each time step $t$, the agent selects an action $a_t \sim \pi(\cdot \mid s_t)$ according to a policy $\pi$, inducing a trajectory $\tau = (s_0, a_0, s_1, a_1, \dots)$. The RL objective is to learn a policy that maximizes the expected discounted return: $
    J(\pi) = \mathbb{E}_{\tau \sim \pi}
    \left[\sum_{t=0}^{\infty} \gamma^t r(s_t, a_t)\right]
$.



\begin{figure}[t]
     \centering
     \begin{subfigure}[b]{0.19\textwidth}
         \centering
         \includegraphics[width=\textwidth]{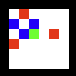}
         \caption{Box Push}
     \end{subfigure}
     \hfill
     \begin{subfigure}[b]{0.19\textwidth}
         \centering
         \includegraphics[width=\textwidth]{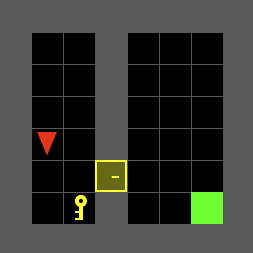}
         \caption{Doorkey}
     \end{subfigure}
     \hfill
     \begin{subfigure}[b]{0.19\textwidth}
         \centering
         \includegraphics[width=\textwidth]{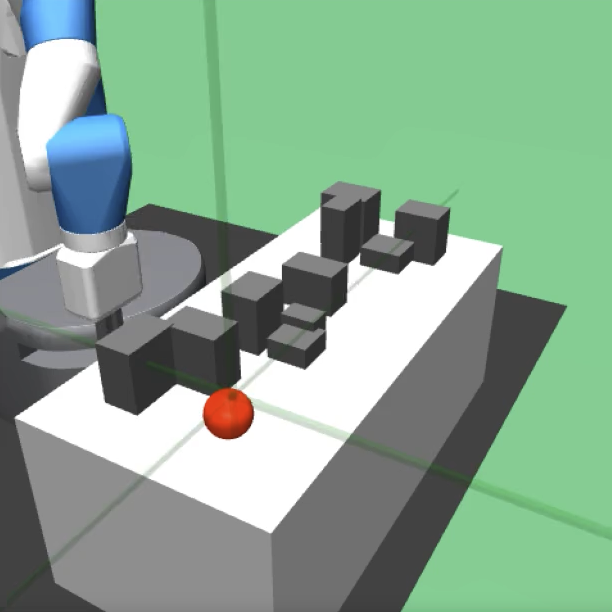}
         \caption{Fetch}
     \end{subfigure}
     \hfill
     \begin{subfigure}[b]{0.19\textwidth}
         \centering
         \includegraphics[width=\textwidth]{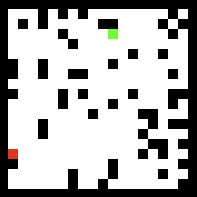}
         \caption{Maze}
     \end{subfigure}
     \hfill
     \begin{subfigure}[b]{0.19\textwidth}
         \centering
         \includegraphics[width=\textwidth]{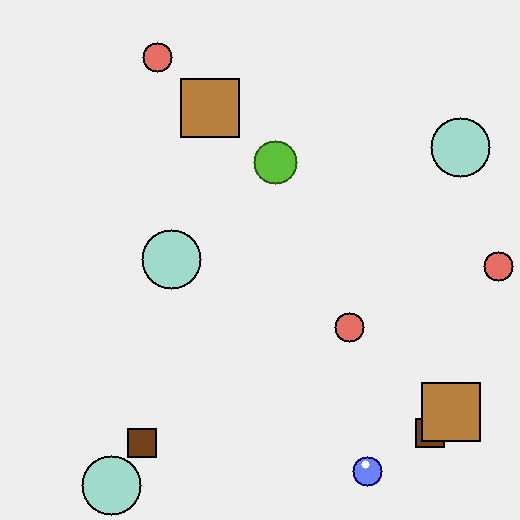}
         \caption{Navigation}
     \end{subfigure}
     \hfill
     \caption{Example images of our environment suite. Environments vary in state space type (discrete versus continuous), task structure, and reactive policy difficulty. In all environments, the objective is to reach a goal state from a start state.}
     \label{fig:env_images}
\end{figure}

\section{Related Work}
\label{sec:related_work}

\paragraph{Learning to use computation.} Several works use RL to learn meta-reasoning. Similar to our work, \citet{callaway_learning_2018} uses model-free RL to select computations. However, their work focuses on allocating compute before taking an action in the real world, while our work frames computation allocation as a sequential decision problem that persists across a full task episode. Other works adapt the amount of compute within a planner \citep{budd_stop_2024, wang_dynamic_2025}, whereas our agent decides between executing a reactive policy instantly or spending time to generate a plan. Event-triggered and adaptive replanning methods \citep{ hashimoto_learning-based_2025, chen_reinforcement_2022, honda_when_2024} learn when a plan is no longer valid, but assume an agent that always plans and does not consider minimizing computation to achieve faster task completion. Other related works consider variable-length options to minimize the number of actions \citep{karimi_dynamic_2023, lakshminarayanan_dynamic_2017, metelli_control_2020, orenstein_toward_2025, sharma_learning_2020}, but these works optimize action efficiency rather than computation allocation, and do not consider an explicit tradeoff between reactive execution and planning cost.

\paragraph{Arbitration between model-free and model-based control.}
The question of when to use model-free versus model-based control has been studied in both AI and neuroscience \citep{dolle_path_2010, lee_neural_2014}. 
Existing AI approaches typically rely on hand-designed switching rules or uncertainty heuristics derived from prediction error signals \citep{bergamaschi_ganapini_fast_2025, sheikhnezhad_fard_novel_2019, fard_mixing_2018, sharma_hybrid_2025, raj_rethinking_2024}, and ignore the computation required for model-based planning. 
Other methods learn to select among multiple controllers \citep{kastner_all--one_2021, dey_learning_2023, hamrick_metacontrol_2017} but assume planning and reactive inference take equal time. 
Our work differs by explicitly modeling the time-based cost of different controllers and learning a meta-reasoning policy to arbitrate between them.

\paragraph{Adaptive computation in LLMs.}
Recent work on LLMs addresses when to generate reasoning tokens \citep{paglieri_learning_2025, fang_thinkless_2025, sabbata_rational_2025, manvi_zero-overhead_2025, chung_thinker_2025}. While structurally similar, these works do not use meta-level agents and instead aim to encode the decision into the LLMs themselves, which constrains the adaptability of the methods. \citet{ong_routellm_2025} uses a controller to route between a larger and smaller model but is trained on preference data instead of learning the decision online as ours does. \citet{hanna_when_2025} introduce a theoretical formalism for when an RL agent should learn to take abstract thinking actions that can be interpreted as choosing additional computation. In contrast, we focus on developing agents that choose from among a set of concrete decision-rules that use varying amounts of computation.

\paragraph{Cognitive science.}
The tradeoff between fast reactive behavior and slow deliberation is well-studied in cognitive science under bounded rationality and dual-process theories \citep{griffiths_doing_2019, keramati_speedaccuracy_2011, kelly_best_2022}. Several works theorize that humans deliberate when the expected value exceeds its cost \citep{lieder_strategy_2017, lieder_rational_2018}, which directly motivates our formulation of planning as a time-costly option. Similar work applies RL as a model of meta-cognitive control \citep{krueger_enhancing_2017, jain_how_2019}, but these works provide descriptive models of human behavior whereas we focus on operationalizing this ability in artificial agents.

\begin{figure}[tbp]
    \centering
    \includegraphics[width=0.98\linewidth]{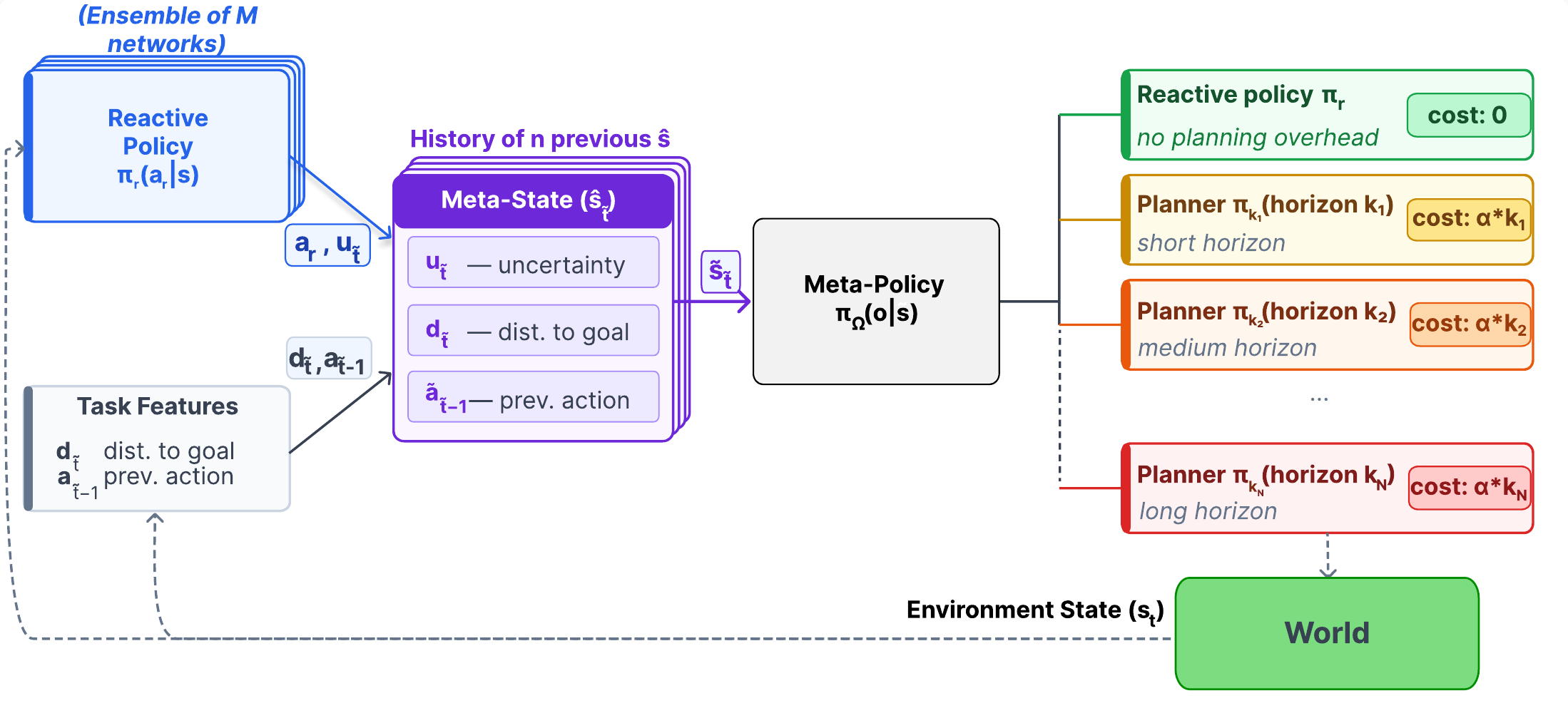}
    \caption{The meta-reasoning agent. The reactive policy (ensemble of neural networks) outputs an action $a_r$ and an uncertainty score $u_{\tilde{t}}$ for the current state. $a_r$ and $u_{\tilde{t}}$, along with task features of distance to goal and previous action, are input to the meta-agent. The meta-agent chooses between a reactive action and a set of plans with different horizons. The actions are executed in the environment.}
    \label{fig:method}
\end{figure}

\section{Problem Formulation}
\label{sec:problem}
 
We consider the problem of training, with RL, a meta-reasoning policy that selects between controllers with different time-based costs.
Specifically, we are interested in switching between a reactive policy, which produces correct actions within its training distribution but costs little time, and a planner, which produces correct actions across the entire state space but costs more time.
We formalize this meta-reasoning problem using the options framework \citep{sutton_between_1999}. The meta-MDP is defined as $\tilde{\mathcal{M}} = (\tilde{\mathcal{S}}, \mathcal{O}, \tilde{\mathcal{P}}, \tilde{r}, \gamma)$ where $\tilde{\mathcal{S}}$ is the meta-state space, $\mathcal{O}$ is the options space, where an option $o \in \mathcal{O}$ is a possibly temporally-extended choice action that executes in the world environment, $\tilde{\mathcal{P}}$ is the meta-transition function, $\tilde{r}$ is the meta-reward and $\gamma$ is the discount factor.
 
The meta-state space $\tilde{\mathcal{S}}$ is the space of the agent's possible internal states, which need not correspond directly to the world-environment state space $\mathcal{S}$.
There are two types of options $\mathcal{O}$ available to the meta-agent: planning options and reactive control options.
We are particularly interested in understanding how the tradeoff between decision speed and decision quality affects task performance.
With this in mind, we restrict ourselves to goal-reaching domains where optimal behavior minimizes time to reach the goal state.
 
Planning options correspond to selecting a planner $\pi_k$ constrained to a fixed planning horizon $k$.
For a plan option of horizon $k$, the planner computes a sequence $p = [a_0, a_1, \dots, a_{k-1}]$ which is then executed to completion in the world environment.
We unify the cost to plan and cost to act under a time-based scale to enable optimizing under the time-to-goal objective.
We define the plan option cost as $c_p(k) = -\alpha k$, where $\alpha$ scales the time-cost per search depth.
Each execution of an action incurs a cost $c_a = -1$ (i.e., takes one unit of time), so the total reward of executing a planning option is $\tilde{r}_{\tilde{t}} = c_p(k) + k \cdot c_a = -\alpha k - k$.
 
Reactive control is modeled as a one-step option where the policy is a learned reactive policy $\pi_{\mathrm{r}}(a \mid s)$. This option incurs no planning cost ($c_p = 0$) and executes a single environment action, therefore $\tilde{r}_{\tilde{t}} = c_a = -1$. The reactive policy is trained, through imitation learning, so that it produces reasonable actions on a limited amount of the full state space. Formally, the reactive policy is trained on a subset of states $\mathcal{S}_{ID} \subset \mathcal{S}$, with the remaining states $\mathcal{S}_{OOD}=\mathcal{S}/\mathcal{S}_{ID}$ considered out-of-distribution. For in-distribution states, the reactive policy performs near-optimally such that $V^{\pi_{\mathrm{r}}}(s) \approx V^*(s)$ for $s \in \mathcal{S}_{ID}$. For out-of-distribution states $s \in \mathcal{S}_{OOD}$, the reactive policy may generalize poorly such that $V^{\pi_{\mathrm{r}}}(s) \ll V^*(s)$, motivating the use of planning despite the incurred computational cost.
 
In the meta-transition dynamics $\tilde{\mathcal{P}}$, transitions between states are induced by the underlying world dynamics and the meta-action chosen. When a planning option of horizon $k$ is taken, the meta-state transition occurs after $k$ world-environment steps. When a reactive control option is chosen, the transition occurs after one world-environment step. The only time a $k$-step planning option terminates early is when the underlying world-environment terminates partially through an option. At each meta-decision timestep $\tilde{t}$, the meta-policy $\pi_{\Omega}(o \mid \tilde{s}_{\tilde{t}})$ selects an option $o$ based on the agent's internal information state $\tilde{s}_{\tilde{t}}$ to control the agent's use of computation. When an option terminates, the meta-policy observes the updated internal state and selects the next option.
 
\section{Reinforcement Learning-Based Meta-Agent for When to Plan}
\label{sec:method}
 
In this section, we describe how to learn a meta-policy for allocating computation between reactive control and decision-time planning.
Our meta-observation design is motivated by three properties of agents operating in large, open-ended environments.
First, as the reactive policy will encounter both in-distribution and out-of-distribution states, the meta-observation cannot use raw environment state, since the meta-policy would then face the same in-distribution/out-of-distribution problem.
Second, the reactive policy may improve over time through experience \citep{anthony2017thinking}, so the meta-observation must reflect current reactive competence.
Third, since the primary benefit of the reactive policy is that it requires no planning time, the meta-observation must itself be computable without search.
 
With these observations in mind, we instantiate the meta-MDP as follows. The meta-policy operates on an internal state $\tilde{s}$ which can be computed without search. The observation consists of an uncertainty score from the reactive policy $u_{\tilde{t}}$, the current distance to the goal $d_{\tilde{t}}$, and the previous meta-action taken $\tilde{a}_{\tilde{t}-1}$. These values are concatenated as a single observation:$$\hat{s}_{\tilde{t}} = [u_{\tilde{t}}, d_{\tilde{t}}, \tilde{a}_{\tilde{t}-1}]$$
Additionally, we provide the meta-agent with a history of $n$ meta-observations $\hat{s}_{\tilde{t}}$, giving it access to trends in uncertainty and progress toward the goal:
 
$$\tilde{s}_{\tilde{t}} = [\hat{s}_{\tilde{t}}, \hat{s}_{\tilde{t}-1}, \dots, \hat{s}_{\tilde{t}-(n-1)}]$$
 
The action space for our meta-agent includes a reactive option and $N$ planning policy options: \[
\mathcal{O} =
\{
\pi_{\mathrm{r}},
\pi_{k_1},
\pi_{k_2},
\dots,
\pi_{k_N}
\}
\]
Each planning option corresponds to a fixed planning horizon, with longer horizons producing longer plans at a greater time cost.
 
A meta-decision is made whenever an option terminates. At each meta-step (\hyperref[fig:method]{Figure \ref{fig:method}}), the reactive policy produces an action and uncertainty value, the meta-state is constructed, and the meta-policy selects an option. If the reactive option $\pi_{\mathrm{r}}$ is chosen, the reactive policy's action is executed for one world step. If a planning option $\pi_k$ is chosen, a plan of horizon $k$ is computed using the planner and then executed open-loop in the world environment for its full duration. The meta-policy $\pi_\Omega$ is trained with model-free RL to minimize time to the goal.
 
To compute the uncertainty score of the reactive policy for the meta-state, we implement our reactive policy as an ensemble of $M$ neural networks \citep{lakshminarayanan_simple_2017}. The networks directly map the environment state $s$ to an action $a$, and using the logits of the output, we compute the uncertainty score for the meta-state. This uncertainty score is a task-agnostic estimate of in-distribution competence based on ensemble disagreement. For continuous control, each network outputs an action mean $\mu_i \in \mathbb{R}^D$. The ensemble action and uncertainty are computed as:
\[
a = \frac{1}{M}\sum_{i=1}^{M} \mu_i, \qquad u(s) = \sum_{j=1}^{D} \mathrm{Var}_i\!\left[\mu_{i,j}\right].
\]
For discrete action spaces, each network outputs a categorical distribution $\mathbf{p}_i$ and the averaged distribution is $\bar{\mathbf{P}} = \frac{1}{M}\sum_{i=1}^{M}\mathbf{p}_i$. The ensemble action and uncertainty are:
\[
a = \arg\max_{j}\, \bar{P}(j), \qquad u(s) = -\sum_{j=1}^{J} \bar{P}(j)\log \bar{P}(j).
\]
In both cases, $u_{\tilde{t}} = u(s_{\tilde{t}})$, and high uncertainty corresponds to out-of-distribution inputs where planning may be beneficial, as the reactive policy is more likely to produce worse actions and therefore take more time to reach the goal.


\section{Experiments}
\label{sec:experiments}

We evaluate our method on a suite of environments in which the goal is to navigate from a starting location to a goal location (\hyperref[fig:env_images]{Figure \ref{fig:env_images}}). First, we show that our meta-policy reaches goal states in fewer timesteps than fixed-compute baselines, including always-reactive and always-planning policies. Second, we ablate the design of the meta-agent's observation space, finding that reactive uncertainty and observation history are the most critical components. Third, we characterize how environment factors including planning cost and reactive competence change the learned allocation strategy. Finally, we show the meta-policy tracks improving reactive competence during joint training, shifting to fully reactive control as both the meta-policy and reactive policy converge.

\subsection{Environments}

Our environment suite (\hyperref[fig:env_images]{Figure \ref{fig:env_images}}) has five environments with varying characteristics. For each environment, we generate distinct sets of object configurations along with a start and goal location, which we consider to be a task. Unless otherwise stated, we use 20 tasks and assign half to reactive in-distribution and half to reactive out-of-distribution. For all environments except Maze, each task has a single start and goal location. For Maze, each task has 5 start and goal configurations.

Here we provide a summary of each environment:
\begin{enumerate}
    \item \textbf{Box Push}: A simplified version of Sokoban, a standard planning benchmark where the goal is to push boxes onto goal locations \citep{botea_using_2003}. In our implementation, the available grid space is 6x6, and there are three boxes. In this domain, an incorrect move (e.g., moving a box into an incorrect corner) can be irrecoverable and cause task failure. The reactive actors receive a symbolic representation of the world.
    \item \textbf{DoorKey}: An environment from the MiniGrid library \citep{MinigridMiniworld23}. DoorKey is a discrete, multi-stage task where the objective is to navigate to a goal position by picking up a key, unlocking a door, and walking to the goal. We modified the environment to use symbolic observations instead of the original image observations. 
    \item \textbf{Fetch}: An environment modified from the Gymnasium Robotics Fetch Reach environment \citep{gymnasium_robotics2023github}. We added obstacles to the table that the gripper must navigate around. The reactive actors receive two images of the state of the world from different angles. The planner for this domain operates in Cartesian space, while the reactive actor operates in 7 DOF joint space to make planning easier while preserving the difficulty of multi-joint arm control. We convert from Cartesian space to joint space through MuJoCo inverse kinematic control for reactive policy training.
    \item \textbf{Maze}: A maze-like gridworld environment where the goal is to navigate to a goal location. Each obstacle configuration defines 5 start/goal pairs to test whether the meta-policy generalizes across targets within the same layout. The reactive actors receive the full maze as an image input.
    \item \textbf{Navigation}: A continuous navigation task. The goal of this environment is to navigate from a start position to a goal position through continuous control. The agent navigates by controlling continuous displacement at each timestep up to a small maximum distance in all directions. The reactive actors receive the full environment as an image input.
\end{enumerate}

\begin{DIFnomarkup}
\begin{table}[t]
\centering
\normalsize
\definecolor{shade}{gray}{0.90} 
\setlength{\tabcolsep}{10pt}
\begin{tabular}{lccccc}
\toprule
\textbf{Environment} & \textbf{Meta-policy} & \textbf{Reactive} & \textbf{Short} & \textbf{Medium} & \textbf{Long} \\
\midrule

\rowcolor{shade} Box Push & $\mathbf{-24.3}$ & $-133.0$ & $-306.1$ & $-44.7$ & $-59.8$ \\
\rowcolor{shade}          & \scriptsize$[0.4, 0.4]$ & \scriptsize$[3.1, 3.1]$ & \scriptsize$[3.6, 3.5]$ & \scriptsize$[2.0, 1.9]$ & \scriptsize$[3.7, 3.6]$ \\
\addlinespace

Doorkey  & $\mathbf{-25.1}$ & $-135.2$ & $-126.7$ & $-27.2$ & $-32.0$ \\
         & \scriptsize$[0.3, 0.3]$ & \scriptsize$[2.5, 2.4]$ & \scriptsize$[2.9, 2.9]$ & \scriptsize$[0.1, 0.1]$ & \scriptsize$[0.1, 0.1]$ \\
\addlinespace

\rowcolor{shade} Fetch    & $\mathbf{-12.2}$ & $-132.4$ & $-51.0$ & $-70.7$ & $-30.0$ \\
\rowcolor{shade}          & \scriptsize$[0.1, 0.1]$ & \scriptsize$[2.4, 2.4]$ & \scriptsize$[2.5, 2.4]$ & \scriptsize$[3.9, 3.7]$ & \scriptsize$[0.3, 0.3]$ \\
\addlinespace

Maze     & $\mathbf{-19.3}$ & $-56.2$  & $-89.6$  & $-37.8$ & $-24.5$ \\
         & \scriptsize$[0.3, 0.3]$ & \scriptsize$[0.9, 0.9]$ & \scriptsize$[1.4, 1.4]$ & \scriptsize$[1.0, 1.0]$ & \scriptsize$[0.4, 0.4]$ \\
\addlinespace

\rowcolor{shade} Navigation & $\mathbf{-39.6}$ & $-140.8$ & $-41.2$ & $-45.4$ & $-46.9$ \\
\rowcolor{shade}            & \scriptsize$[0.1, 0.1]$ & \scriptsize$[2.3, 2.3]$ & \scriptsize$[0.1, 0.1]$ & \scriptsize$[0.1, 0.1]$ & \scriptsize$[0.2, 0.2]$ \\

\bottomrule
\end{tabular}
\caption{Average episode return, which includes both the cost of acting and planning. Baselines are fixed policies that exclusively call the reactive, short plan, medium plan, and long plan options. 95\% bootstrap confidence intervals (margins shown below the mean). Bold entries overlap in CI with the best performer per environment. We run 30 seeds per experiment with 300 evaluation episodes per seed.}
\label{tab:main_results}
\end{table}
\end{DIFnomarkup}

\subsection{Experimental Setup}

\paragraph{Meta-Policy.} We train the meta-policy with PPO \citep{schulman_proximal_2017} implemented in Stable Baselines3 \citep{stable-baselines3} with default hyperparameters. We chose PPO for its stability and well-understood training dynamics. We use a history of 4 observations to capture the short-term trends in reactive uncertainty and task progress.

\paragraph{Reactive Policy.} The reactive policy is pretrained by behavior cloning from trajectories generated by the planner on in-distribution tasks. All experiments use $M=4$ networks. For pretraining parameters see \hyperref[tab:combined_hyperparams]{Appendix \ref{tab:combined_hyperparams}}.

\paragraph{Planning.} Planning is performed using a given accurate world model which isolates the compute allocation problem from the model learning problem. In all environments we use an A* planner which we constrain to a maximum search depth to approximate a time limit\footnote{Several proxies for a time-based budget are possible. In this work, we assign cost based on search depth. An alternative is an effort-based approximation, such as assigning the cost to be the number of A* nodes expanded. Both are proxies for the same underlying time-based cost.} (\hyperref[tab:combined_hyperparams]{Table \ref{tab:combined_hyperparams}}).

For continuous control domains, we use a lattice-based A* over a discretized Euclidean state space --- end-effector space for Fetch and agent location space for Navigation. We use short, medium, and long planning horizons of 15\%, 30\%, and 50\% of the task horizon to span a low, medium, and high amount of compute. We leave the sensitivity of the meta-policy to these values for future work. 

We use a plan cost of $\alpha=0.5$ as the default value. This value was chosen to avoid planning that is prohibitively expensive or trivially cheap. A value of $\alpha=0.5$ intuitively means that each planning step costs 50\% more time units than a reactive step and avoids the two non-interesting regimes. As $\alpha \rightarrow 0$, planning is effectively free, so an optimal policy always plans, and when $\alpha$ is large, planning is dominated by reactive control in all but the most extreme out-of-distribution cases. We analyze the sensitivity of the meta-policy to $\alpha$ in \hyperref[sec:env_characteristics]{Section \ref{sec:env_characteristics}}.

\paragraph{Evaluation.}
We define return to be the total cost of time to plan and act over an episode as stated in \hyperref[sec:problem]{Section \ref{sec:problem}}. We run all experiments over 30 independent seeds and results are aggregated across seeds. All methods are evaluated for 300 episodes using deterministic evaluation after training, and we fix the hyperparameters across all ablations.

\begin{DIFnomarkup}
\begin{table}[t]
\centering
\normalsize
\definecolor{shade}{gray}{0.90} 
\setlength{\tabcolsep}{8pt}
\begin{tabular}{lcccc}
\toprule
\textbf{Environment} & \textbf{Reactive} & \textbf{Short} & \textbf{Medium} & \textbf{Long} \\
\midrule

\rowcolor{shade} Box Push & $44.4$ & $0.0$ & $0.0$ & $55.6$ \\
\rowcolor{shade}  & \scriptsize$[0.8,\,0.8]$ & \scriptsize$[0.0,\,0.0]$ & \scriptsize$[0.0,\,0.0]$ & \scriptsize$[0.8,\,0.8]$ \\
\addlinespace

DoorKey & $23.8$ & $35.4$ & $40.8$ & $0.0$ \\
 & \scriptsize$[0.6,\,0.7]$ & \scriptsize$[0.3,\,0.3]$ & \scriptsize$[0.4,\,0.4]$ & \scriptsize$[0.0,\,0.0]$ \\
\addlinespace

\rowcolor{shade} Fetch & $77.0$ & $20.7$ & $0.0$ & $2.3$ \\
\rowcolor{shade}  & \scriptsize$[0.5,\,0.5]$ & \scriptsize$[0.5,\,0.5]$ & \scriptsize$[0.0,\,0.0]$ & \scriptsize$[0.2,\,0.2]$ \\
\addlinespace

Maze & $51.1$ & $0.3$ & $0.7$ & $47.9$ \\
 & \scriptsize$[1.0,\,1.0]$ & \scriptsize$[0.1,\,0.1]$ & \scriptsize$[0.1,\,0.1]$ & \scriptsize$[1.0,\,1.0]$ \\
\addlinespace

\rowcolor{shade} Navigation & $20.0$ & $80.0$ & $0.0$ & $0.0$ \\
\rowcolor{shade}  & \scriptsize$[0.7,\,0.7]$ & \scriptsize$[0.7,\,0.7]$ & \scriptsize$[0.0,\,0.0]$ & \scriptsize$[0.0,\,0.0]$ \\

\bottomrule
\end{tabular}

\caption{Meta-action percentages for the trained meta-policy with 95\% bootstrap confidence intervals
         (margins shown below each mean). We run 30 seeds per experiment with 300 evaluation episodes per seed.}
\label{tab:action_percentages}
\end{table}
\end{DIFnomarkup}

\subsection{Learning When to Plan}

In this experiment, we verify that our meta-policy learns to effectively allocate compute across our suite of environments. \hyperref[tab:main_results]{Table \ref{tab:main_results}} shows the mean evaluation return for our meta-agent in comparison to fixed-compute baselines, and \hyperref[tab:action_percentages]{Table \ref{tab:action_percentages}} shows the meta-action distribution of the learned meta-policy. In all five domains, our meta-agent performs better than all fixed-compute strategies, showing our adaptive computation method is able to learn effective planning-reacting behavior. The closest heuristic to our meta-agent is the Medium Plan in the Doorkey environment. Notably, Doorkey is also the only domain where the meta-policy does not converge to a two-action strategy. Since Doorkey is a staged task, these results suggest that the medium and short planning horizons align with the sub-goal structure of the task and a fixed strategy can achieve close performance to an adaptive one.

The two tasks where the meta-agent achieves the largest percent improvement over baselines are Box Push with almost 2x improvement and Fetch with about a 2.5x improvement. In Box Push, short-horizon planning is suboptimal in most situations since low-depth search is inadequate to find the long-horizon action sequences required to reach the goal state, as seen in the large performance gap between short and medium planning results. Furthermore, Box Push is the only domain where mistakes are unrecoverable, meaning imprecise use of the reactive policy risks failure. The meta-agent's performance suggests it learns to avoid reactive control and short plans in states where mistakes would cause failure. In Fetch, the complexity stems from the action space of the 7 DOF robot arm the reactive policy must control. Since the meta-agent converges to a predominantly reactive strategy with over 75\% reactive control, it suggests that the reactive policy partially generalizes to out-of-distribution tasks, but cannot complete them without planning. The strong performance in both domains suggests that task complexity does not make the meta-reasoning problem harder. 

In three of the five tasks, the meta-policy converges to a near-equal split between reactive and planning actions, suggesting the agent learns to distinguish between in-distribution and out-of-distribution states. Navigation and Fetch are two exceptions, which likely reflect the reactive policy's out-of-distribution performance. Qualitatively, the navigation reactive policy sometimes gets "stuck" on objects in the task, and the results suggest that the meta-policy can distinguish when that happened and switch to planning.

\begin{DIFnomarkup}
\begin{table}[!t]
\centering
\normalsize
\definecolor{shade}{gray}{0.90}
\setlength{\tabcolsep}{8pt}
\begin{tabular}{lccccc}
\toprule
\textbf{Environment} & \textbf{Meta-policy} & \textbf{Distance} & \textbf{History} & \textbf{Previous Action} & \textbf{Uncertainty} \\
\midrule

\rowcolor{shade} Box Push & $-24.3$ & $\mathbf{-20.9}$ & $-110.7$ & $-36.3$ & $-73.9$ \\
\rowcolor{shade}  & \scriptsize$[0.4,\,0.4]$ & \scriptsize$[0.3,\,0.2]$ & \scriptsize$[2.5,\,2.5]$ & \scriptsize$[1.2,\,1.1]$ & \scriptsize$[2.4,\,2.2]$ \\
\addlinespace

DoorKey & $\mathbf{-25.1}$ & $-34.3$ & $-26.1$ & $-29.1$ & $-27.9$ \\
 & \scriptsize$[0.3,\,0.3]$ & \scriptsize$[1.0,\,0.9]$ & \scriptsize$[0.1,\,0.1]$ & \scriptsize$[0.6,\,0.6]$ & \scriptsize$[0.6,\,0.6]$ \\
\addlinespace

\rowcolor{shade} Fetch & $\mathbf{-12.2}$ & $-12.5$ & $-21.1$ & $\mathbf{-12.1}$ & $-16.1$ \\
\rowcolor{shade}  & \scriptsize$[0.1,\,0.1]$ & \scriptsize$[0.1,\,0.1]$ & \scriptsize$[0.8,\,0.8]$ & \scriptsize$[0.1,\,0.1]$ & \scriptsize$[0.2,\,0.2]$ \\
\addlinespace

Maze & $\mathbf{-19.3}$ & $\mathbf{-19.7}$ & $\mathbf{-19.1}$ & $\mathbf{-19.3}$ & $-25.0$ \\
 & \scriptsize$[0.4,\,0.3]$ & \scriptsize$[0.4,\,0.4]$ & \scriptsize$[0.3,\,0.3]$ & \scriptsize$[0.4,\,0.4]$ & \scriptsize$[0.4,\,0.4]$ \\
\addlinespace

\rowcolor{shade} Navigation & $-39.6$ & $-41.2$ & $-40.7$ & $\mathbf{-37.9}$ & $-41.2$ \\
\rowcolor{shade}  & \scriptsize$[0.1,\,0.1]$ & \scriptsize$[0.1,\,0.1]$ & \scriptsize$[0.1,\,0.1]$ & \scriptsize$[0.2,\,0.2]$ & \scriptsize$[0.1,\,0.1]$ \\

\midrule
Average & $\mathbf{-24.1}$ & $-25.7$ & $-43.5$ & $-27.0$ & $-36.8$ \\
 & \scriptsize$[0.1,\,0.1]$ & \scriptsize$[0.2,\,0.2]$ & \scriptsize$[0.5,\,0.5]$ & \scriptsize$[0.3,\,0.3]$ & \scriptsize$[0.5,\,0.5]$ \\
\bottomrule
\end{tabular}
\caption{Average episode return per method ablation with 95\% bootstrap confidence intervals (margins below each mean). Bold entries overlap in CI with the best performer per environment. Each header is the component of the ablation we remove from the observation space (e.g., \textbf{Distance} removes the distance to goal). We run 30 seeds per experiment with 300 evaluation episodes per seed.}
\label{tab:method_ablation}
\end{table}
\end{DIFnomarkup}

\subsection{Ablation Study}
\label{sec:ablation_study}

In this section, we investigate which components of our meta-agent's observation space are most critical for learning adaptive computation. We ablate each component by removing one feature at a time:
\begin{enumerate}
    \item \textbf{Distance}: Removes the distance component.
    \item \textbf{History}: Meta-agent only receives current state observations.
    \item \textbf{Previous Action}: Removes the previous action taken.
    \item \textbf{Uncertainty}: Removes the model-free uncertainty.
\end{enumerate}

The results in \hyperref[tab:method_ablation]{Table \ref{tab:method_ablation}} show that the two most important components of our method are the history and uncertainty scores. Poor performance without History is most clearly seen in the Box Push (4.5x worse) and Fetch  (1.7x worse) environments. We once again note that these tasks have challenges in irreversibility of actions and high-dimensional action space, respectively. By removing history, it seems as if the agent is no longer able to track progress, and therefore it is unclear if the reactive agent is moving toward the goal. A similar drop-off is seen without the uncertainty score in these two environments. This suggests that without directly being able to tell if the task is in-distribution or out-of-distribution for the reactive policy, the meta-agent is not able to perform optimally. However, the lack of substantial drop-off without the uncertainty score in the other three environments suggests that the combination of history and distance enables the agent to learn a proxy for uncertainty in these environments, but that learned metric is not as effective as directly accessing the uncertainty.

\subsection{Environment Characteristics}
\label{sec:env_characteristics}

\begin{figure}[t]
    \centering
    \includegraphics[width=0.95\linewidth]{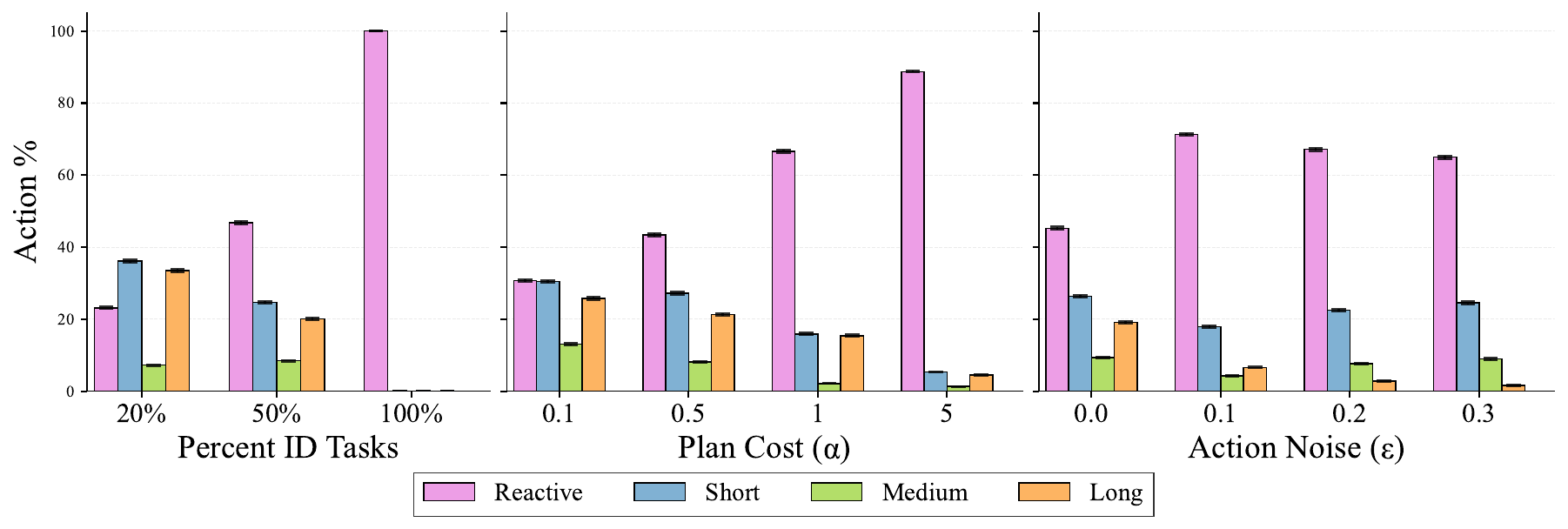}
    \caption{Average meta-action taken after training across all environments for different environment characteristics. We vary the percentage of in-distribution tasks (\textbf{left}), the computational cost of planning (\textbf{center}), and environmental stochasticity (\textbf{right}). Results reflect 95\% bootstrap confidence intervals across 30 seeds per experiment (300 evaluation episodes per seed).}
    \label{fig:env_experiment}
\end{figure}
    
In this section, we investigate how different environment characteristics affect the learned meta-policy behavior. First, we vary the proportion of in-distribution tasks by changing the number of out-of-distribution tasks. Results are shown in \hyperref[fig:env_experiment]{Figure \ref{fig:env_experiment}} in the leftmost section. As the proportion of in-distribution tasks increases, the meta-policy shifts toward reactive control, with the proportion of reactive actions roughly tracking the proportion of in-distribution tasks. This behavior is expected since the reactive policy is competent in a higher proportion of tasks, so the meta-policy correctly identifies that planning is less frequently necessary. At the extreme where all tasks are in-distribution, the meta-policy converges to fully reactive control, confirming the meta-policy correctly interprets the reactive policy's competence across the distribution of tasks.

Second, we vary the planning cost parameter $\alpha$, where a value of $\alpha=0.1$ means a plan of length 15 costs 1.5 timesteps to compute. Results are shown in \hyperref[fig:env_experiment]{Figure \ref{fig:env_experiment}} in the middle section. As planning cost increases, the meta-policy relies increasingly on reactive control. Notably, even though the reactive policy produces suboptimal actions in out-of-distribution states, the time saved by avoiding planning is sufficient to achieve better returns than planning in some cases. A second finding is that the ratio of short, medium, and long planning horizons remains roughly constant as $\alpha$ increases, suggesting the meta-policy reduces planning frequency uniformly rather than exclusively dropping longer plans. This implies the meta-policy learns when to plan, but not how long to plan, as the distribution of plan lengths remains consistent across environments.

Third, we vary the stochasticity of the environment by introducing action noise $\epsilon$ into the transition dynamics. For discrete environments, $\epsilon$ controls the probability of a random action being taken, and for continuous environments, we add Gaussian noise with variance $\epsilon$ to the action. Results are shown in \hyperref[fig:env_experiment]{Figure \ref{fig:env_experiment}} in the rightmost section. The most significant shift occurs between no noise and low noise ($\epsilon=0.1$), where the reactive action percentage increases from approximately 45\% to 70\%. The increase in reactive action percentage is driven mostly by a reduction in long planning, which drops from roughly 20\% to 5\%. This result is consistent with the intuition that longer plans are most affected by increased noise since fewer replanning actions mean errors compound more over the plan duration. As noise increases further, this trend continues, with long planning decreasing with each increase in $\epsilon$. Interestingly, as noise increases from $\epsilon=0.1$ to $\epsilon=0.3$, the reactive percentage decreases slightly while short and medium planning increase slightly. This result suggests that while the reactive policy handles high noise better than long planning, short and medium plans retain utility in out-of-distribution states since half of the tasks remain out-of-distribution for the reactive policy. The meta-policy therefore converges to a mixed strategy that favors reactive control more than the baseline $\epsilon=0$ but sometimes uses short planning for out-of-distribution states under high stochasticity.

\begin{figure}[t]
    \centering
    \includegraphics[width=0.95\textwidth]{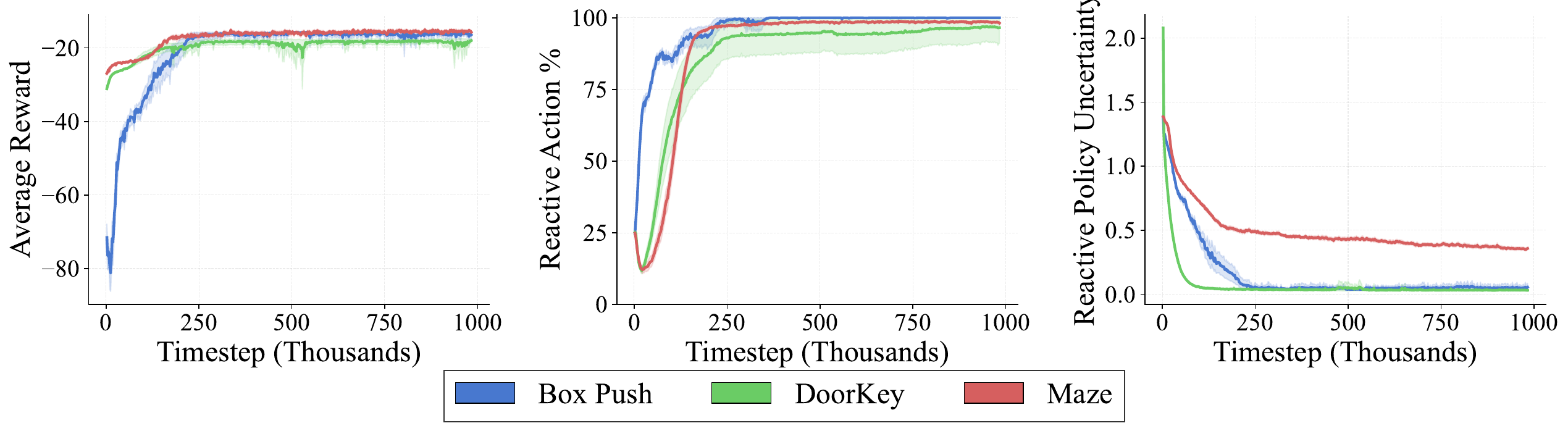} 
    \caption{ Average return, reactive action percent, and reactive actor uncertainty value as the meta-policy is trained. The reactive policy is trained on a history buffer of planner actions after every rollout. 30 seeds per environment. Shaded regions indicate the 95\% confidence interval. Box Push, DoorKey, and Maze only; Fetch and Navigation excluded due to compute constraints.}
    \label{fig:joint_training_plot}
\end{figure}

\subsection{Joint-Training Setting}
\label{sec:joint_training}

In this section, we empirically evaluate the ability of our method to adapt in a joint-training setting where the reactive actor improves alongside the meta-agent. To train the reactive policy, we record the state-action pairs produced by planning options and add these to a replay buffer. After each meta-agent rollout, all reactive ensemble members are trained by behavior cloning on this buffer. As joint training requires significantly longer training runs, we restrict this experiment to Box Push, DoorKey, and Maze due to compute constraints. Training hyperparameters are reported in \hyperref[tab:combined_hyperparams]{Table \ref{tab:combined_hyperparams}}.

Results are shown in \hyperref[fig:joint_training_plot]{Figure \ref{fig:joint_training_plot}}. As training progresses, the meta-policy shifts toward reactive control across all environments, converging to fully reactive execution as the reactive actor improves. This is also seen in the reactive uncertainty signal, which converges to low disagreement across environments and confirms the ensemble uncertainty score reflects the strength of the reactive policy. Together, these results show the meta-policy is able to adapt as the reactive policy learns and converges to the optimal strategy.

An interesting property of this training setup is that the reactive policy is trained on data generated from when a planning option is called. This structure creates a learning curriculum where the meta-agent must call planning in out-of-distribution states, which simultaneously produces the best immediate reward and generates the training data for the reactive policy. Once the reactive policy has been trained on out-of-distribution states, the meta-policy then recognizes that the reactive policy uncertainty has decreased, revealing that the reactive policy is now competent in more states and stops calling planning options. This shift maximizes the return and avoids overtraining the reactive policy on already in-distribution states.

\section{Discussion and Future Work}
\label{sec:discussion}

Our empirical results demonstrate that RL-based meta-policies can successfully learn adaptive computation allocation from uncertainty signals and high-level state features across a suite of motion planning and navigation tasks. 
However, several assumptions we make point toward natural future work.
First, our method assumes access to a perfect world model for planning, which does not hold in most real-world settings.
Cognitive science work on meta-reasoning suggests that humans account for model uncertainty when deciding whether to plan \citep{glascher_states_2010}, so extending our method to incorporate learned world models with state uncertainty is a natural next step.

Second, we constrain our environments to be single-agent and static during plan creation. Widening the environment suite to include tasks without such constraints and enabling the meta-agent to interrupt executing plans would test whether the reactive-planning tradeoff we study can generalize beyond the synchronous and single-agent setting.

Third, our meta-observation relies on hand-designed features. While our ablations confirm these are effective, more expressive uncertainty estimates (e.g., Bayesian or flow-based methods) or learned observation representations could improve meta-policy generalization, particularly in high-dimensional observation spaces such as images.

Our joint-training setup also suggests a promising direction for future work.
Our meta-agent was able to track an improving reactive policy over the course of training so an interesting question is whether it can also track a \emph{worsening} reactive policy.
This is a setting that may occur when the reactive policy has insufficient capacity to represent optimal actions in all states across a big world \citep{javed_big_nodate} or suffers from catastrophic forgetting.

More broadly, the reactive-planning tradeoff is applicable to many adaptive-computation problems. In LLM settings, the tradeoff between fast single-pass generation and slower chain-of-thought reasoning is structurally similar to the reactive-planning tradeoff we study in this work. Some VLA works \citep{team_gemini_2025} adopt a similar structure with a fast and lightweight model on a robot and a larger but slower model in the cloud. Our uncertainty-conditioned meta-policy provides a method for learning these allocations.

\section{Conclusion}
\label{sec:conclusion}

In this work, we presented an RL method for learning adaptive computation allocation through a meta-reasoning policy. Our approach conditions on reactive policy uncertainty and high-level state features to select between a fast reactive option and planning options, where planning improves action quality at the cost of time. Across a suite of motion planning and navigation tasks, our meta-policy outperforms all fixed-compute heuristic baselines when optimizing for a time-to-goal objective. Ablations confirm that reactive uncertainty and observation history are the most critical components of the meta-state. We further showed that the learned meta-policy adapts to environment characteristics: increasing planning cost shifts allocation toward reactive control, and increasing reactive in-distribution coverage shifts allocation away from planning. Finally, we demonstrated that conditioning on uncertainty signals enables the meta-policy to converge to exclusively reactive control in a joint-training setting where the reactive policy continues to learn. Together, these results show that high-level state and reactive competence signals are sufficient for a meta-reasoning policy to effectively allocate computation across environments and training regimes. 


\appendix

\subsubsection*{Acknowledgments}
\label{sec:ack}
We thank Brahma Pavse and the RLC reviewers for providing helpful feedback which improved this work before and during the review process. This work took place in the Prediction and Action Lab (PAL) at the University of Wisconsin–Madison. PAL research is supported by NSF (IIS-2410981) and the Wisconsin Alumni Research Foundation. Additionally, Adam Labiosa is supported by an NSF Research Traineeship award, No. 2152163.


\bibliography{references}
\bibliographystyle{rlj}

\beginSupplementaryMaterials

%




\section{Appendix 1: Hyperparameters}
\label{sec:hyperparams}

\begin{DIFnomarkup}
\begin{table}[h]
\centering
\small
\begin{tabular}{lccccc}
\toprule
& \textbf{Box Push} & \textbf{DoorKey} & \textbf{Fetch} & \textbf{Maze} & \textbf{Navigation} \\
\midrule
\multicolumn{6}{l}{\textbf{Reactive Pretraining}} \\
Steps collected & 10,000 & 10,000 & 2,000 & 10,000 & 10,000 \\
Epochs & 30 & 30 & 30 & 30 & 30 \\
Loss & CE & CE & MSE & CE & MSE \\
\midrule
\multicolumn{6}{l}{\textbf{Planner}} \\
Task horizon & 30 & 25 & 30 & 12 & 30 \\
Short / Med / Long plan & 5/10/15 & 4/8/12 & 5/10/15 & 2/4/6 & 5/10/15 \\
\bottomrule
\end{tabular}
\caption{Environment-specific hyperparameters for reactive pretraining and planner configuration. CE = Cross-Entropy, MSE = Mean Squared Error. Batch size 64 across all environments.}
\label{tab:combined_hyperparams}
\end{table}
\end{DIFnomarkup}

\begin{DIFnomarkup}
\begin{table}[h]
\centering
\small
\begin{tabular}{ll}
\toprule
\textbf{Hyperparameter} & \textbf{Value} \\
\midrule
\multicolumn{2}{l}{\textbf{Joint-Training Reactive Policy}} \\
Train frequency & 2048 steps \\
Buffer size & 10,000 \\
Epochs / Batch size & 3 / 64 \\
Learning rate & 1e-4 \\
\bottomrule
\end{tabular}
\caption{Joint-training hyperparameters.}
\label{tab:joint_training_hyperparams}
\end{table}
\end{DIFnomarkup}

\section{Reactive Policy Architecture}

Here we detail the architectures used for the reactive policy.

\begin{DIFnomarkup}
\begin{table}[H]
\centering
\small
\setlength{\tabcolsep}{4pt} 
\begin{tabular}{lllc}
\toprule
\textbf{Actor Type} & \textbf{Obs. Type} & \textbf{Action} & \textbf{Architecture} \\
\midrule
\textbf{Image} & Image & Discrete & 
    \begin{tabular}[t]{@{}l@{}}
      \textit{CNN:} Conv(C$\to$16, 3$\times$3) $\to$ ReLU $\to$ MP(2) \\
      \textit{MLP:} FC($n\to$512) $\to$ ReLU $\to$ FC(512$\to d$)
    \end{tabular} \\
\addlinespace[8pt]

\textbf{Vector} & Tensor & Discrete & 
    \begin{tabular}[t]{@{}l@{}}
      \textit{MLP:} FC($d_{\text{in}}\to$512) $\to$ ReLU $\to$ FC(512$\to$512) \\
      $\to$ ReLU $\to$ FC(512$\to d$)
    \end{tabular} \\
\addlinespace[8pt]

\textbf{Image-Cont} & Image & Continuous & 
    \begin{tabular}[t]{@{}l@{}}
      \textit{CNN:} [Conv $\to$ ReLU $\to$ MP(2)]$\times$3 $\to$ AAP(8$\times$8) \\
      \textit{Shared:} FC($n\to$1024) $\to$ ReLU $\to$ FC(1024$\to$512) \\
      \textit{Head:} FC(512$\to a$) $\to$ Tanh
    \end{tabular} \\
\bottomrule
\end{tabular}
\caption{Model-free actor network architectures. Image is used for Maze. Vector is used for BoxPush and DoorKey. Image-Cont is used for Fetch and Navigation. MP\,=\,MaxPool2d, AAP\,=\,AdaptiveAvgPool2d, FC\,=\,fully-connected, $d$\,=\,\texttt{features\_dim}, $a$\,=\,\texttt{action\_dim}, $n$\,=\,CNN output size.}
\label{tab:model_free_actors}
\end{table}
\end{DIFnomarkup}

\section{Separated Environment Ablations}

Here we show the results in \hyperref[fig:env_experiment]{Figure \ref{fig:env_experiment}} non-averaged over environments.

\begin{figure}[H] 
    \centering
    \includegraphics[width=0.95\linewidth]{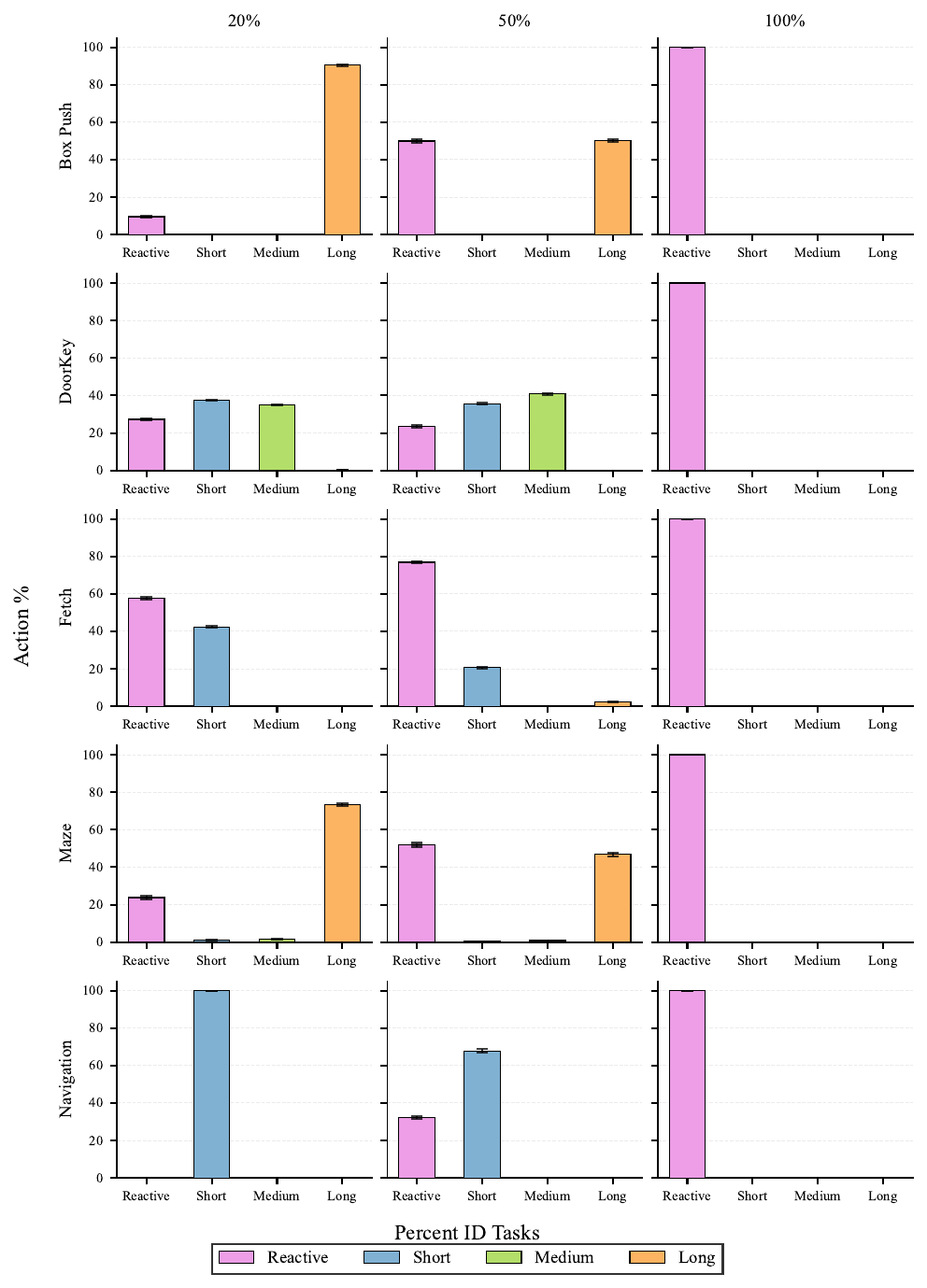} 
    \caption{\textbf{Reactive policy in-distribution percentage experiment.} Individual environment results when varying the percent of in-distribution tasks for the reactive policy.}
    \label{fig:supplemental_indist}
\end{figure}

\begin{figure}[H] 
    \centering
    \includegraphics[width=0.95\linewidth]{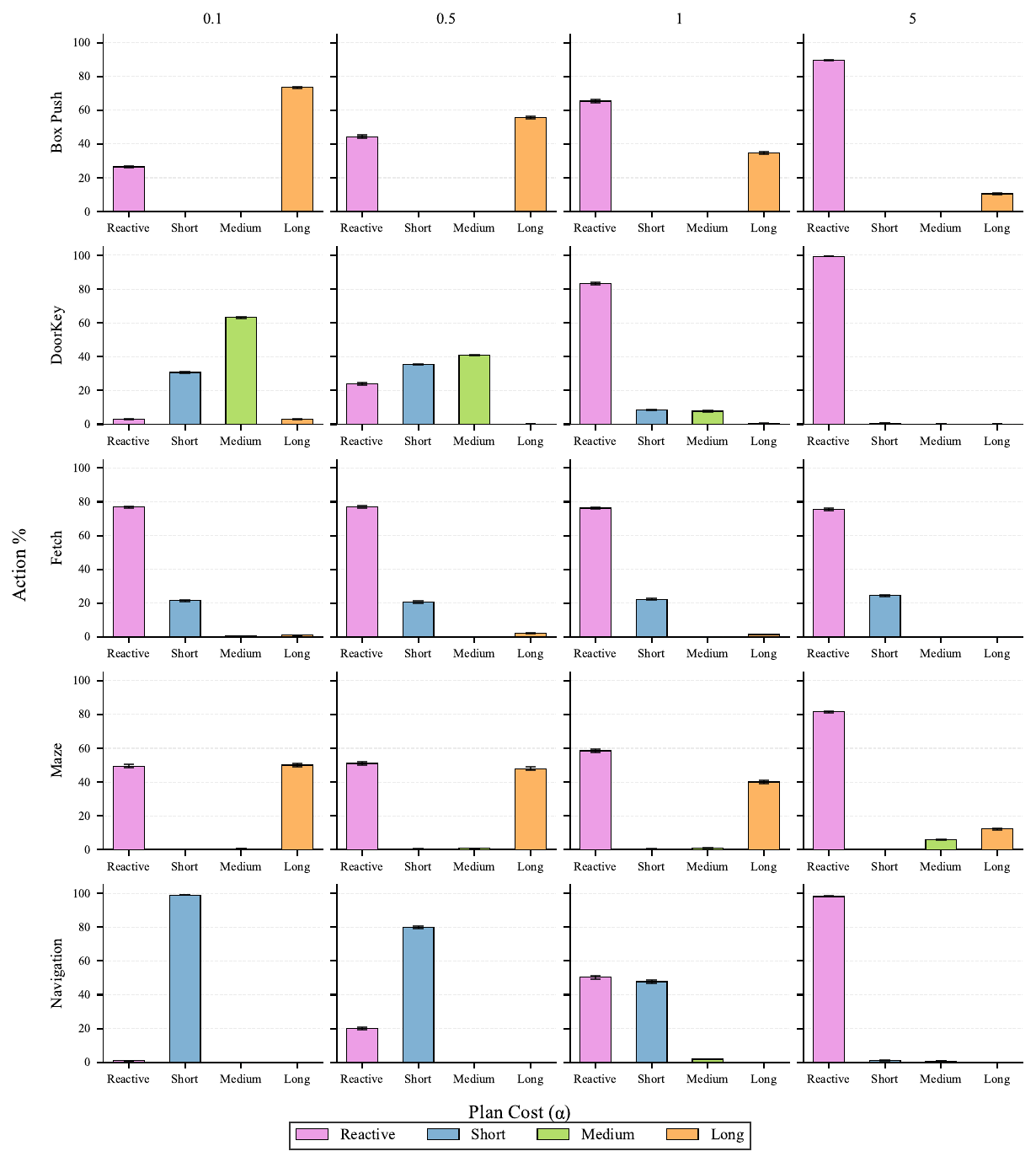} 
    \caption{\textbf{Plan cost experiment.} Individual environment results when varying the planning cost $\alpha$.}
    \label{fig:supplemental_plancost}
\end{figure}

\begin{figure}[H] 
    \centering
    \includegraphics[width=0.95\linewidth]{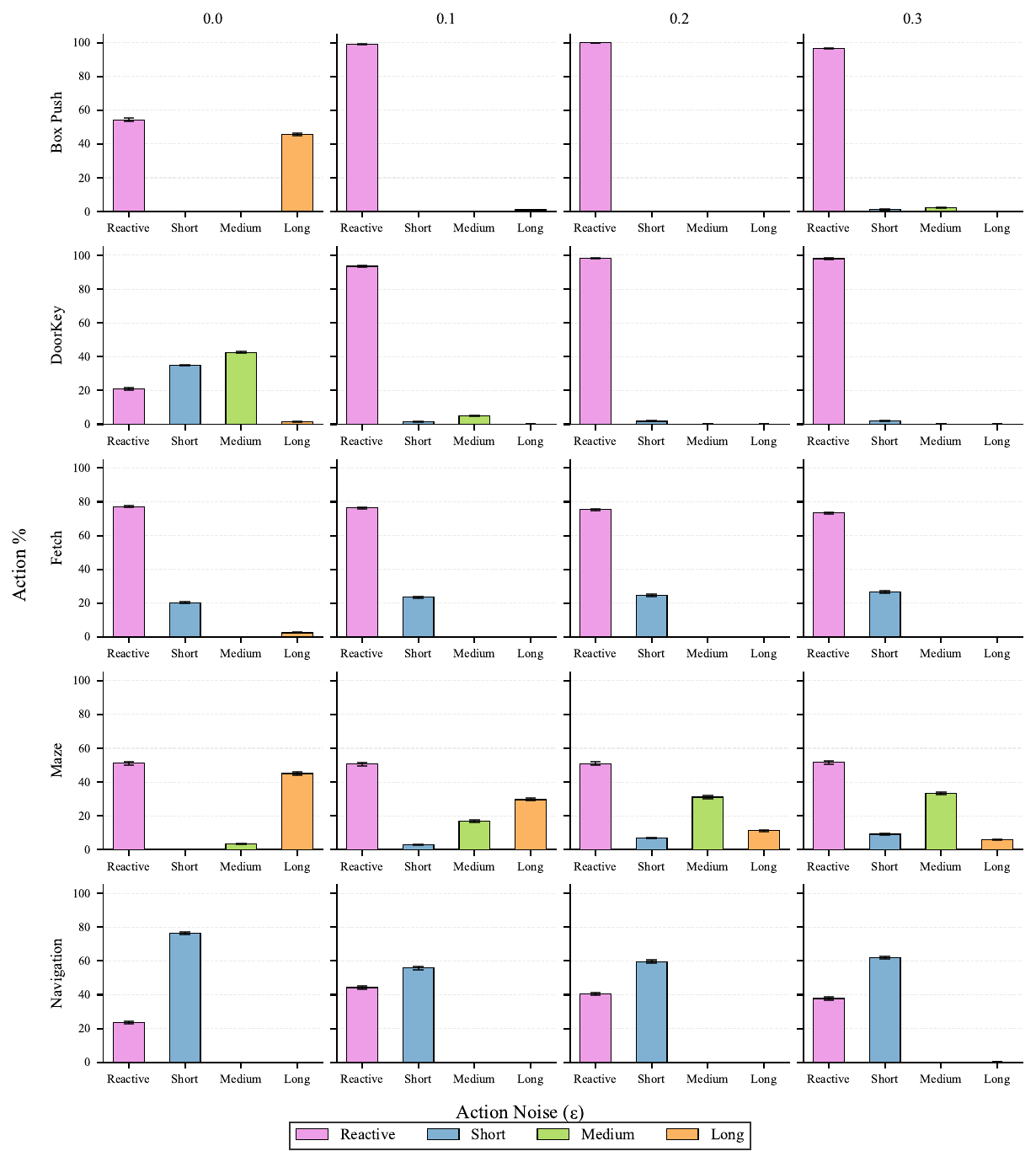} 
    \caption{\textbf{Action stochasticity experiments.} Individual environment results when varying the environment stochasticity.}
    \label{fig:supplemental_noise}
\end{figure}

\end{document}